\documentclass[a4paper,11pt]{article}

\usepackage{graphicx}  %
\usepackage{url}       %
\usepackage{times}
\usepackage{natbib}
\usepackage[margin=25mm]{geometry}

\usepackage{color}
\usepackage{amsmath}
\usepackage{amssymb}
\usepackage{tikz}
\usepackage{wrapfig}
\usepackage{graphicx}
\usepackage{subfig}
\usepackage{multirow}
\usepackage{multicol}
\usepackage{booktabs}
\usepackage{enumitem}

\newcounter{tbsnr}
\newenvironment{tbs}
{\addtocounter{tbsnr}{1}\par\bigskip\noindent\fbox{\thetbsnr}
\hspace*{\fill}\begin{minipage}{7cm}\tt}
{\end{minipage}\hspace*{\fill}\bigskip}

\newcommand{\cut}[1]{}

\title{Evaluating the Representational Hub\\
of Language and Vision Models}
\date{}

\author{Ravi Shekhar$^\dagger$, Ece Takmaz$^*$, Raquel Fern\'{a}ndez$^*$ and Raffaella Bernardi$^\dagger$ \\
$^\dagger$University of Trento, $^*$University of Amsterdam  \\
{\tt raffaella.bernardi@unitn.it} \ \ \  {\tt raquel.fernandez@uva.nl}
}

\begin{document}
\maketitle
\thispagestyle{empty}
\pagestyle{empty}

\begin{abstract}
\noindent
The multimodal models used in the emerging field at the intersection
of computational linguistics and computer vision implement the
bottom-up processing of the ``Hub and Spoke'' architecture proposed in
cognitive science to represent how the brain processes and combines
multi-sensory inputs. In particular, the Hub is implemented as a
neural network encoder. We investigate the effect on this encoder of
various vision-and-language tasks proposed in the literature: visual
question answering, visual reference resolution, and visually grounded
dialogue.  To measure the quality of the representations learned by
the encoder, we use two kinds of analyses. First, we evaluate the
encoder pre-trained on the different vision-and-language tasks on an
existing {\em diagnostic task} designed to assess multimodal semantic
understanding. Second, we carry out a battery of analyses aimed at
studying how the encoder merges and exploits the two
modalities. %

\end{abstract}

\section{Introduction}
\label{sec:intro}

In recent years, a lot of progress has been made within the emerging field at 
the intersection of computational linguistics and computer vision thanks to the use of 
deep neural networks. 
The most common strategy to move the field forward has been to propose different multimodal tasks---such as visual question answering~\citep{anto:vqa15}, visual question generation~\citep{gene:most16}, visual reference resolution~\citep{KazemzadehOrdonezMattenBergEMNLP14}, and visual dialogue~\citep{visdial}---and to develop task-specific models. 

The benchmarks developed so far have put forward complex and distinct
neural architectures, but in general they all share a common backbone
consisting of an encoder which learns to merge the two types of
representation to perform a certain task.  This resembles the
bottom-up processing in the `Hub and Spoke' model proposed in
Cognitive Science to represent how the brain processes and combines
multi-sensory inputs~\citep{patt:theh15}.  In this model, a `hub'
module merges the input processed by the sensor-specific `spokes' into
a joint representation. We focus our attention on the encoder
implementing the `hub' in artificial multimodal systems, with the goal
of assessing its ability to compute multimodal
representations that are useful beyond specific tasks.

While current visually grounded models perform remarkably well on the
task they have been trained for, it is unclear whether they are able
to learn representations that truly merge the two modalities and
whether the skill they have acquired is stable enough to be
transferred to other tasks. In this paper, we investigate these
questions in detail. To do so, we evaluate an encoder trained on
different multimodal tasks on an existing {\em diagnostic task}---FOIL
\citep{shekhar2017foil_acl}---designed to assess multimodal semantic
understanding and carry out an in-depth analysis to study how the
encoder merges and exploits the two modalities. 
We also exploit two techniques to investigate the structure of the 
learned semantic spaces: 
Representation Similarity
Analysis (RSA)~\citep{kriegeskorte2008representational} and 
Nearest Neighbour overlap (NN). We use
RSA to compare the outcome of the various encoders given the same
vision-and-language input and NN to compare the multimodal space 
produced by an encoder with
the ones built with the input visual and language embeddings, respectively,
which allows us to measure the relative weight an encoder gives to the two
modalities.

In particular, we consider three visually grounded tasks: visual question
answering (VQA)~\citep{anto:vqa15}, where the encoder is trained to
answer a question about an image; visual resolution of referring
expressions (ReferIt)~\citep{KazemzadehOrdonezMattenBergEMNLP14}, where 
 the model has to pick up the referent object of a description in an image; and GuessWhat~\citep{guesswhat_game}, where the model
has to identify the object in an image that is the target of a
goal-oriented question-answer dialogue. %
We make
sure the datasets used in the pre-training phase are as similar as
possible in terms of size and image complexity, and use the same model
architecture for the three pre-training tasks. This guarantees fair
comparisons and the reliability of the results we obtain.\footnote{The datasets are
  available at \url{https://foilunitn.github.io/}.}

We show that the multimodal encoding skills learned by pre-training the
model on GuessWhat and ReferIt are more stable and transferable than
the ones learned through VQA. This is reflected in the lower number of
epochs and the smaller training data size they need to reach their
best performance on the FOIL task. 
We also observe that the semantic spaces learned by the encoders trained
on the ReferIt and GuessWhat tasks are closer to each other than to 
the semantic space learned by the VQA encoder. Despite these asymmetries
among tasks, we find that all encoders give more weight to the visual input than the linguistic one.

\section{Related Work}
\label{sec:related}

Our work is part of a recent research trend that aims at analyzing, interpreting, and evaluating neural models 
by means of auxiliary tasks besides the task they have been trained for~\citep{adi2016fine,linzen2016assessing,alishahi-etl-conll2017,zhan:lang18,conn:what18}.
Within  language and vision research, the growing interest in having a better understanding of what neural models really learn has led to the creation of several diagnostic
datasets~\citep{girs:clev16,shekhar2017foil_acl,suhr-EtAl:2017:Short}.

Another research direction which is relevant to our work is transfer learning,
a machine learning area that studies how the skills learned by a model
trained on a particular task can be transferred to learn a new task
better, faster, or with less data. Transfer learning has proved
successful in computer vision %
(e.g.~\cite{raza:Cnnf14}) 
as well as in computational linguistics (e.g.,~\cite{cann:super17}).
However, little has been done in this respect for visually grounded
natural language processing models. 

In this work, we combine these different research lines and explore transfer learning techniques in the domain of language and vision tasks. In particular, we use the FOIL diagnostic dataset~\citep{shekhar2017foil_acl} 
and investigate to what extent skills learned through different multimodal tasks transfer. 

While transfering the knowledge learned by a pre-trained model can be useful in principle, \cite{conn:what18} found that randomly initialized models provide
strong baselines that can even outperfom pre-trained
classifiers (see also~\cite{wieting2019no}).
However, it has also been shown that these untrained, randomly initialized
models can be more sensitive to the size of the training set
than pre-trained models are~\citep{zhan:lang18}. We will investigate
these issues in our experiments.

\section{Visually Grounded Tasks and Diagnostic Task}
\label{sec:tasks}

We study three visually grounded tasks:~visual question answering (VQA), visual resolution of referring expressions (ReferIt), and goal-oriented dialogue for visual target identification (GuessWhat). While ReferIt was originally
formulated as an object detection task~\citep{KazemzadehOrdonezMattenBergEMNLP14}, VQA~\citep{anto:vqa15} and GuessWhat~\citep{guesswhat_game} were defined as
classification tasks. Here we operationalize
the three tasks as retrieval tasks, which makes comparability easier.

\begin{itemize}[itemsep=0pt]
\item \textbf{VQA:} Given an image and a natural language question about it, the model is
trained to retrieve the correct natural language answer out of a list of possible
answers.

\item \textbf{ReferIt:} Given an image and a natural language description of an entity in
the image, the model is asked to retrieve the bounding box of the
corresponding entity out of a list of candidate bounding boxes. 

\item \textbf{GuessWhat:} Given an image and a natural language
question-answer dialogue about a target entity in the image, the model
is asked to retrieve the bounding box of the target among a list of
candidate bounding boxes. The GuessWhat game also involves asking questions before guessing. Here we focus on the guessing task that takes place after the question generation step.
\end{itemize}

\begin{figure}\centering
\begin{minipage}{3.3cm}
\includegraphics[width=3.1cm]{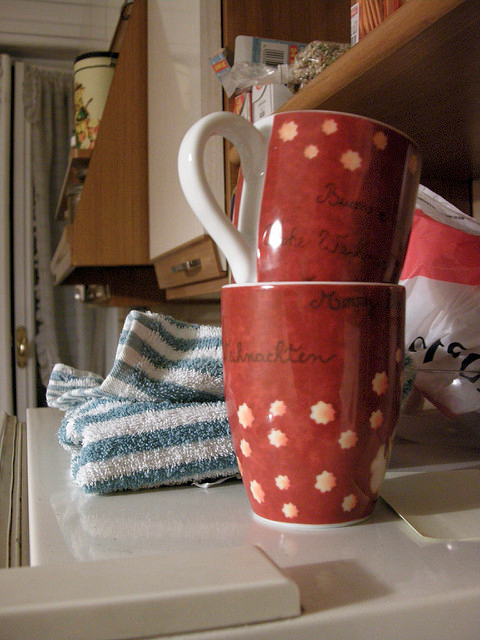}
\end{minipage}
\begin{minipage}{4.8cm}\footnotesize
\textbf{VQA}\\
{\em Q: How many cups are there? \\ A: Two.}

\vspace{5pt}
\textbf{ReferIt}\\
{\em The top mug.}

\vspace{5pt}
\textbf{GuessWhat}\\
{\em Q: Is it a mug?\\
A: Yes\\
Q: Can you see the cup's handle?\\
A: Yes.}
\end{minipage}
\begin{minipage}{3.2cm}
\includegraphics[width=3.1cm]{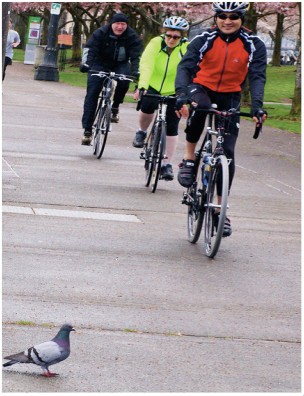}
\end{minipage}
\begin{minipage}{3.8cm}\footnotesize
\textbf{FOIL Diagnostic Task}\\

\textbf{original caption}\\
{\em Bikers approaching a bird.}\\

\textbf{foiled caption}\\
{\em Bikers approaching a \underline{dog}.}

\vspace{1.3cm}
\end{minipage}
\caption{Illustrations of the three visually-grounded tasks (left) and
  the diagnostic task (right).
}\label{fig:example}
\end{figure}

\noindent
Figure~\ref{fig:example} (left) exemplifies the similarities and differences among the three tasks. All three tasks require merging and encoding visual and linguistic input. In VQA, the system is trained to make a language-related prediction, while in ReferIt it is trained to make visual predictions. GuessWhat includes elements of both VQA and ReferIt, as well as specific properties: The system is trained to make a visual prediction (as in ReferIt) and it is exposed to questions (as in VQA); but in this case the linguistic input is a coherent sequence of visually grounded questions and answers that follow a goal-oriented strategy and that have been produced in an interactive setting.

To evaluate the multimodal representations learned by the encoders of the models trained on each of the three tasks above, we leverage the FOIL task (concretely, task 1 introduced by~\cite{shekhar2017foil_acl}), a binary classification task designed to detect semantic incongruence in visually grounded language.

\begin{itemize}[itemsep=0pt]
\item \textbf{FOIL (diagnostic task):} 
Given an image and a natural language caption describing it, the model is asked to decide
whether the caption faithfully describes the image or not, 
i.e., whether it contains a foiled word that is incompatible with the
image (foil caption) or not (original caption). 
Figure~\ref{fig:example} (right) shows an example in
  which the foiled word is ``dog''.
Solving this task requires  some degree of compositional alignment between modalities, which is key for fine-grained visually grounded semantics.

\end{itemize}

\section{Model Architecture and Training}
\label{sec:models}

In cognitive science, the hub module of~\cite{patt:theh15} receives representations
processed by sensory-specific spokes and computes a multimodal
representation out of them. 
All our models have a common core that resembles this
architecture, while incorporating some task-specific components. 
This allows us to investigate the impact of specific tasks 
on the multimodal representations computed by the representational hub, 
which is implemented as an encoder.
Figure~\ref{fig:modelarchitecture} shows a diagram of the shared model components, 
which we explain in detail below.
 
\subsection{Shared components}
\label{sec:shared}
To facilitate the comparison of the representations learned via the 
different tasks we consider, 
we use pre-trained visual and linguistic features
to process the input given to the encoders. 
This provides a common initial base across models and diminishes the effects of using
different datasets for each specific task (the datasets are described in Section~\ref{sec:setup}). 

\paragraph{Visual and language embeddings}  
To represent visual data, we use ResNet152
features~\citep{he2016:resnet}, which yield state of the art performance in image
classification tasks and can be computed efficiently. 
To represent linguistic data, we use Universal
Sentence Encoder (USE) vectors~\citep{cer2018universal} since 
they yield near state-of-the-art results 
on several NLP tasks and are suitable both for short texts (such as
the descriptions in ReferIt) and longer ones (such as the dialogues in
GuessWhat).\footnote{The dialogues in the GuessWhat?!~dataset consist
  of 4.93 question-answer pairs on average \citep{guesswhat_game}.}

In order to gain some insight into the semantic spaces that emerge from these visual and linguistic representations, 
we consider a sample of 5K datapoints sharing the images across the three tasks and use average cosine similarity as a
measure of space density. We find that 
the semantic space of the input images is denser (0.57 average cosine similarity) than the semantic space of the linguistic input across all tasks (average cosine similarity of 0.26 among VQA questions, 0.35 among ReferIt descriptions, and 0.49 among GuessWhat dialogues). 
However, when we consider the retrieval candidates rather than the input data, we find a different pattern:
The linguistic semantic space of the candidate answers in VQA is much denser than the visual space of the candidate bounding boxes in ReferIt and GuessWhat (0.93 vs.~0.64 average cosine similarity, respectively).
This suggests that the VQA task is harder, since the candidate answers are all highly similar.

\begin{figure}
\begin{center}
\includegraphics[width=13cm]{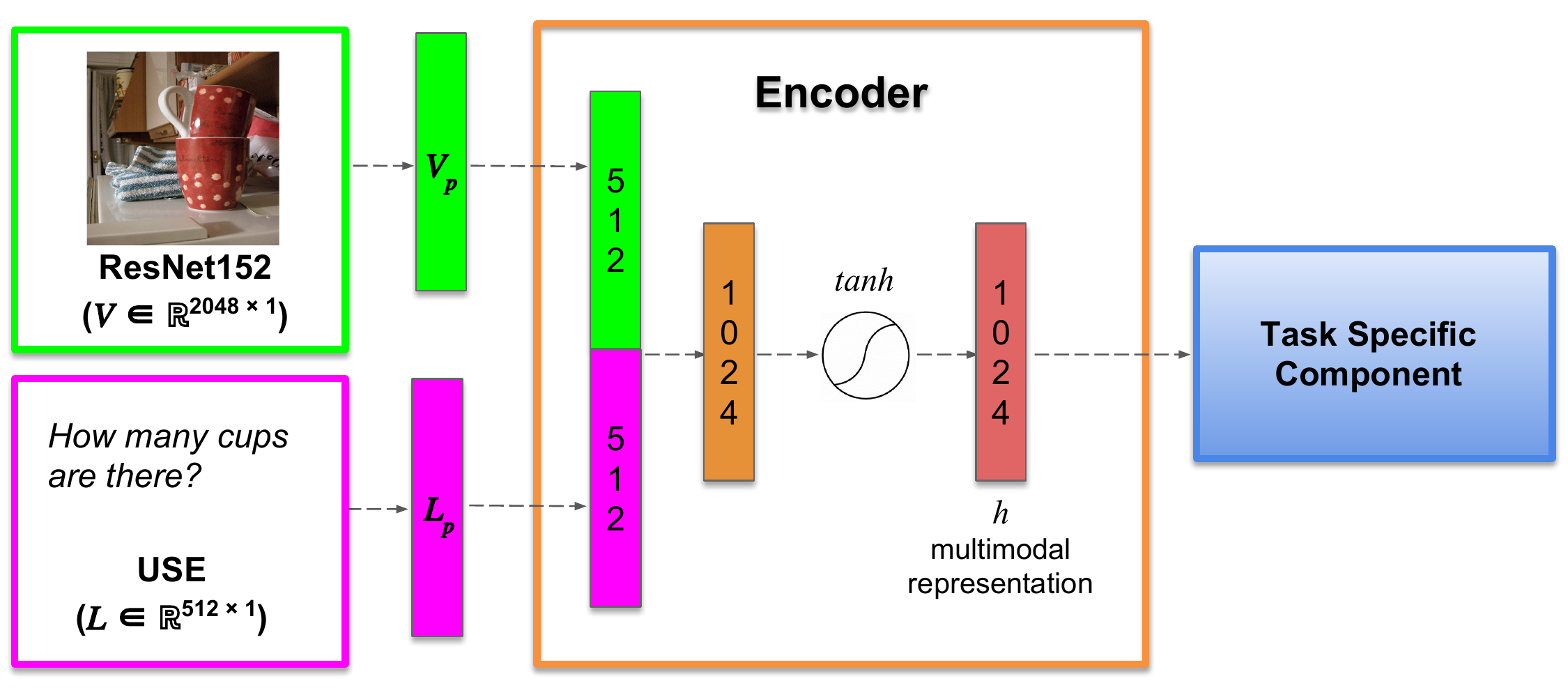} %
\end{center}
\caption{General model architecture, with an example from VQA as input. The encoder receives as input visual (ResNet152) and linguistic (USE) embeddings and merges them into a multimodal representation ($h$). This is passed on to a task-specific
  component: an MLP in the case of the pre-training retrieval tasks and a fully connected layer in the case of the FOIL classification task.
\label{fig:modelarchitecture}}
\end{figure}

\paragraph{Encoder} 
As shown in Figure~\ref{fig:modelarchitecture}, 
ResNet152 visual features ($V \in \mathbb{R}^{2048 \times 1}$)
and  USE linguistic features ($L \in \mathbb{R}^{512 \times 1}$) are input in the model and passed
through fully connected layers that project them onto spaces of
the same dimensionality. 
The projected representations ($V_{p}$ and
$L_{p}$) are concatenated,
passed through a linear layer, and then through a \emph{tanh}
activation function, which produces the final encoder representation $h$:
\begin{equation}%
h = \tanh \left( W \cdot \left[V_{p};\ L_{p}\right] \right)
\end{equation}
where $W \in \mathbb{R}^{1024 \times 1024}$, $V_{p} \ \in \mathbb{R}^{512 \times 1}$, $L_{p} \in \mathbb{R}^{512 \times 1}$, and $[\cdot ; \cdot]$ represents concatenation.

\subsection{Task-specific components}
\label{sec:task-specific}

The architecture described above is shared by all the models
we experiment with, which thus differ only with respect
to their task-specific component.
 
\paragraph{Pre-training task component} 
For the three tasks we consider, the final encoder representation $h$
is given to a Multi-Layer Perceptron (MLP), which generates either a
language embedding (VQA model) or a visual embedding (ReferIt and
GuessWhat models).  The three task-specific models are
  trained with a cosine similarity loss, which aims to get the
  generated embedding closer to the ground truth embedding and farther
  away from any other embeddings in the list of candidates. Details of
  how, for each datapoint, the list of candidates  is selected are provided
  in Section~\ref{sec:setup}. The embeddings of such candidates are
  obtained with USE (for VQA) and ResNet (for ReferIt and
  GuessWhat). As mentioned above, the high density of the VQA candidate answers' space
  makes the task rather hard.

\paragraph{FOIL task component} 
To evaluate the encoder representations learned by the pre-trained models, 
the task-specific MLPs are replaced by a fully
connected layer, which is trained on the FOIL task using a cross-entropy loss. 
We train the FOIL task component using the following settings:

\begin{itemize}[itemsep=0pt]

\item \textbf{Random$_2$} The encoder weights are randomly initialized
 and the FOIL classifier layer is untrained. 
 This provides a lower-bound baseline with random performance.
 
\item \textbf{Random} The encoder weights are randomly initialized
 and then frozen
  while the FOIL classifier layer is trained on the FOIL task. 
  This provides a strong baseline that is directly comparable to the task-specific setting explained next.

\item \textbf{Pre-trained (VQA, ReferIt, GuessWhat)} The encoder weights are initialized with the Random setting's seeds and the model is trained on each of the
tasks. The weights of the task-specific encoders are then frozen and the
  FOIL classifier layer is trained on the FOIL task. With this
  setting, we are able to diagnose the transfer and encoding properties
  of the pre-trained tasks.  

\item \textbf{Fully trained on FOIL} The encoder weights are initialized with the
  Radom setting's seeds. Then the full model is trained on the FOIL task, 
  updating the weights of the projected vision and language layers, the encoder, 
  and the FOIL layer. 
  This provides the upper
  bound on the FOIL classification performance, as the entire model is optimized for this task from the start.

\end{itemize}

\section{Experimental Setup}
\label{sec:setup}

We provide details on the data sets and the 
implementation settings we use in our experiments. 

\paragraph{Pre-training datasets} 
For the three visually grounded tasks, we use the
VQA.v1 dataset by~\cite{anto:vqa15}, the RefCOCO dataset by
\cite{yu2016modeling}, and the GuessWhat?!~dataset by
\cite{guesswhat_game} as our starting point.
All these datasets have been developed with images from
MS-COCO~\citep{lin:micr14}. 
We construct {\em common image} datasets
for by taking the intersection of the images in
the  three original datasets. This results in a total of 14,458 images. 
An image can be part of several data points, i.e, it can be paired with  more than 
one linguistic input. Indeed, the 14,458 common images correspond to
43,374 questions for the VQA task, 104,227 descriptions for the ReferIt task, and 35,467 dialogues for the GuessWhat task. 

To obtain datasets of equal size per task that are as similar as possible, we filter the resulting data points according to the following procedure: 

\begin{enumerate}[itemsep=0pt]
\item For each image, we check how many
linguistic items are present in the three datasets and fix the
minimum number ($k$) to be our target number of linguistic items paired with that image.
\item We select $n$ data points where the descriptions in ReferIt and dialogues in GuessWhat concern the same target object (with $n \leq k$). 
\item Among the $n$ data points selected in the previous step, we select the $m$ data points in VQA where the question or the answer mention the same target object (computed by string matching). 
\item We make sure all the images in each task-specific dataset are paired with exactly $k$ linguistic items; if not, we select additional ones randomly until this holds. 
\end{enumerate}

\noindent
This results in a total of 30,316 data points per dataset: 14,458 images shared across datasets, paired with 30,313 linguistic items. 
We randomly divided this {\em common image} dataset into
training and validation sets at the image level. The training set
consists of 13,058 images (paired with 27,374 linguistic items) 
and the validation set of 1,400 images (paired with 2,942 linguistic items). 
Table~\ref{tab:datasetsta} provides an overview of the datasets.

\begin{table*}[t]
\begin{center}
\begin{tabular}{l|cc|ccc}\toprule
& \multicolumn{2}{|c|}{\bf common image datasets} & \multicolumn{3}{|c}{\bf FOIL dataset} \\
 & training & validation & training & validation & testing \\\hline
\# images & 13,058 & 1,400 & 63,240 & 13,485 &20,105 \\
\# language & 27,374 & 2,942 &  358,182 & 37,394  & 126,232\\\bottomrule
\end{tabular}
\caption{Statistics of the datasets used for the pre-training tasks and the FOIL task.}\label{tab:datasetsta}
\end{center}
\end{table*}

As mentioned in Section~\ref{sec:tasks}, we operationalize the three
tasks as retrieval tasks where the goal is to retrieve the correct
item out of a set of candidates. 
In the VQA.v1 dataset (multiple choice version), there are 18 candidate answers 
per question. In GuessWhat?!~there are on average 18.71 candidate objects per dialogue, all of them appearing in the image. We take the same list of candidate objects per image for the ReferIt task.

\paragraph{FOIL dataset} The FOIL dataset consists of image-caption 
pairs from MS-COCO and pairs where the caption has been modified
by replacing a noun in the original caption with a foiled noun, such that
the foiled caption is incongruent with the image---see Figure~\ref{fig:example} for an example and \cite{shekhar2017foil_acl} for further details on the construction of the 
dataset.\footnote{\cite{madhysastha-wang-specia_NAACL:2018} found that an earlier version
of the FOIL dataset was biased. We have used the latest version of the dataset available at 
\url{https://foilunitn.github.io/}, which does not have this problem.}
The dataset contains 521,808 %
captions (358,182 in training, 37,394 in validation and 126,232 in test set) and %
96,830 images (63,240, 13,485 and 20,105, in training, validation and test set,
respectively) -- see Table~\ref{tab:datasetsta}. All the images in the test set do not occur either in
the FOIL training and validation set, nor in the common image dataset
described above and used to pre-train the models.

\paragraph{Implementation details}
All models are trained using supervised learning with ground truth data. 
We use the same parameters for all models: batch size of 256 and Adam optimizer~\citep{kingma2014:adam} with learning rate 0.0001. 
All the parameters are tuned on the validation set. Early stopping is used while training, i.e., training is stopped when there is no improvement on the validation loss for 10 consecutive epochs or a maximum of 100 epochs, and the best model is taken based on the validation loss.

\section{Results and Analysis}

We carry out two main blocks of analyses: one exploiting FOIL as
diagnostic task and the other one investigating the structure of the semantic spaces 
produced by the pre-trained encoders when receiving the same multimodal
inputs.

Before diving into the results of these analyses, 
we evaluate the three task-specific models on the tasks they have been trained for.
Since these are retrieval tasks, we compute precision at rank 1 (P@1) on the validation sets and compare the results to chance performance. Given the number of candidate answers and objects per task in our datasets, chance P@1 is 0.055 for VQA and 0.05 for ReferIt and GuessWhat. Our task-specific models obtain P@1 values of 0.14 for VQA (mean rank 2.84), 
0.12 for ReferIt (mean rank 3.32), and 
0.08 for GuessWhat (mean rank 4.14).
Not surprisingly given the challenging nature of these tasks, the results are not high. Nevertheless, the representations learned by the models allow them to perform above chance level and thus provide a reasonable basis for further investigation.

\subsection{Analysis via diagnostic task}

In this first analysis, we assess the quality of the multimodal representations learned by the three multimodal tasks considered in terms of their potential to perform the FOIL task, i.e., to spot semantic (in)congruence between an image and a caption. Besides comparing the models with respect to task accuracy, we also investigate how they learn to adapt to the FOIL task over training epochs, how much data they need to reach their best performance, and how confident they are about the decisions they make.

\paragraph{FOIL accuracy} 

Table~\ref{tab:accuracy} shows accuracy results on the FOIL  task for the different training settings described in Section~\ref{sec:task-specific}.
We report accuracy for the task overall, as well as accuracy on detecting original and foiled
captions. 
As expected, the Random$_2$ setting yields chance performance ($\approx$50\% overall, with a surprisingly strong preference for classifying captions as foiled). 
The model fully trained on FOIL achieves an accuracy of 67.59\%. 
This confirms that the FOIL task is challenging, as shown by \cite{shekhar2017foil_acl}, even for models that are optimized to solve it. 
The Random setting, where a randomly initialized encoder is trained on the FOIL task, yields 53.79\% accuracy overall -- higher than the chance lower bound by Random$_2$, but well below the upper bound set by the fully trained model.

The key results of interest for our purposes in this paper are those achieved by the models
where the encoder has been pre-trained on each of the three multimodal tasks we study. 
We observe that, like the Random encoder, the pre-trained encoders achieve results well below the upper bound.
The VQA encoder yields results comparable to Random, while ReferIt and GuessWhat achieve slightly higher results: 54.02\% and 54.18\%, respectively. 
This trend is much more noticeable when we zoom into the accuracy results on original vs.~foiled captions. 
All models (except Random$_2$) achieve lower accuracy on the foil class than on the original class. However, the GuessWhat encoder performs substantially better than the rest: Its foil accuracy is not only well above the Random encoder, but also around 2\% points over the fully trained model (49.34\% vs.~47.52\%). The ReferIt encoder also performs reasonably well (on a par with the fully trained model), while the VQA encoder is closer to Random.

This suggests that the ReferIt and the GuessWhat encoders do learn a small degree of multimodal understanding skills that can transfer to new tasks. The VQA encoder, in contrast, seems to lack this ability by and large. 

\begin{table}[!ht]
\begin{center}
\begin{tabular}{l|c|ccc|}\toprule
 & \bf overall & \bf original & \bf foiled\\ \midrule
Random$_2$ & 49.99 & 0.282   & 99.71 \\ %
Random &53.79 & 65.33 & 42.25 \\ \hline
VQA  &53.78 & 66.09  & 41.48 \\
ReferIt &54.02& 60.39 & 47.66\\
GuessWhat &54.18& 59.02 & 49.34\\\hline
Fully FOIL &67.59& 87.66 &47.52\\ \bottomrule
\end{tabular}
\end{center}
\caption{Accuracy on the FOIL task for the best model of each training setting.}\label{tab:accuracy}
\end{table}

\begin{figure*}
\centering  \hspace*{-8.5pt}
\subfloat[Training epochs.\label{fig:epochs}]
	{{\includegraphics[height=3.87cm]{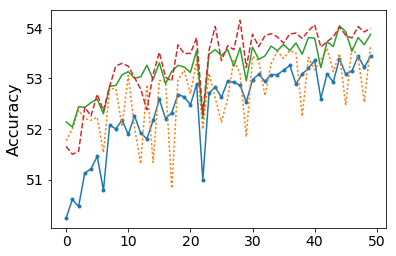} }} \hspace*{-8.5pt}
\subfloat[Size of FOIL training set (log scaled).\label{fig:size}]
	{{\includegraphics[height=3.87cm]{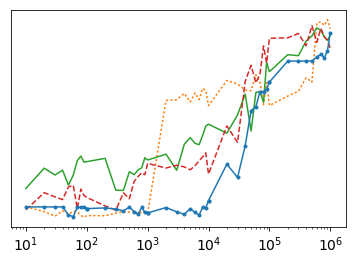}}}\hspace*{-6pt}
\subfloat[AUC indicating confidence.\label{fig:auc}]
	{{\includegraphics[height=3.87cm]{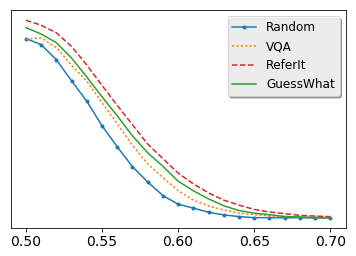}}}
\caption{Comparisons among the pre-trained encoders and the randomly initialized encoder, regarding their accuracy over training epochs, with varying data size, and across different decision thresholds.} 
\end{figure*}

\paragraph{Learning over time} 
In order to better understand the effect
of the representations learned by the pre-trained encoders, we trace the evolution of the FOIL
classification accuracy over time, i.e., over the first 50 training epochs. %
As shown in Figure~\ref{fig:epochs}, all the pre-trained models start with
higher accuracy than the Random model. This shows that the encoder is
able to transfer knowledge from the pre-trained tasks to some
extent. The Random model takes around 10 epochs to catch up and after that
it does not manage to improve much. 
The evolution of the accuracy achieved by the ReferIt and GuessWhat encoders
is relatively smooth, i.e., it increases progressively with further 
 training epochs. The one by the VQA model, in contrast, is far less stable.
    
\paragraph{Size of FOIL training data} 
Next, we evaluate how the accuracy achieved by the models changes when varying the 
size of the FOIL training set. By controlling the amount of training data,
we can better tease apart whether the performance of the pre-trained models is 
due to the quality of the encoder representations or simply
to the amount of training the models undergo on the FOIL task itself.
Figure~\ref{fig:size} gives an overview.  The GuessWhat encoder has a clear advantage 
when very little training data is available, while the other encoders start at chance level. 
Both GuessWhat and ReferIt increase their accuracy relatively smoothly as more data is
provided, while for the VQA model there is a big jump in accuracy once enough FOIL data is available. 
Again, this suggests that the representations learned by the GuessWhat encoder are of 
somewhat higher quality, with more transferable potential.

\paragraph{Confidence} 
Finally, we analyse the confidence of the models by measuring their
Area Under the Curve (AUC). We gradually increase
the classification threshold from 0.5 to 0.7 by an interval of 0.01. This measures the
confidence of the classifier in making a prediction. As shown in
Figure~\ref{fig:auc}, all models have rather low confidence (when the threshold
is 0.7 they are all at chance level). 
The Random model exhibits the lowest confidence, while the ReferIt model
is slightly more confident in its decisions than the rest, followed by the GuessWhat model.

\subsection{Analysis of the multimodal semantic spaces learned by the encoders}
In this section, we analyse the encoders by comparing the similarity of
the multimodal spaces they learn
and by comparing the learned multimodal spaces to the visual and linguistic 
representations they receive as input in terms on nearest neighbours.

\paragraph{Representation similarity analysis} 
Representation
Similarity Analysis (RSA) is
a technique from neuroscience~\citep{kriegeskorte2008representational} that has
been recently leveraged in computational linguistics, for example to compare the semantic
spaces learned by artificial communicating
agents~\citep{bouchacourt2018agents}.
It compares different semantic spaces by comparing their internal similarity relations, given a common set $N$ of input data points.
Each input $k \in N$ is processed by an encoder for a given task $Ti$, 
producing vector $h_{Ti}^k$. 
Let $H_{Ti}^N$ be the set of vector representations created by the encoder of $Ti$ for all the items in $N$; and let $H_{Tj}^N$ be the corresponding set of representations by the encoder of task $Tj$. 
These two semantic spaces, $H_{Ti}^N$ and $H_{Tj}^N$, are not directly comparable as they have been produced independently.
RSA remedies this by instead comparing their structure in terms of internal similarity relations. 
By computing cosine similarity between all pairs of vectors within each semantic space, we obtain a vector of cosine similarities per space, 
which captures its internal structure. 
These similarity vectors have identical dimensionality, namely $N(N-1)/2)$ values, and hence can be directly compared by computing Spearman correlation between them. 
The resulting RSA scores (corresponding to the aforementioned Spearman correlation coefficients) tell us the extent to which the two
 sets of representations are structurally similar.

\begin{figure}[t]
\centering
\includegraphics[width=0.4\linewidth]{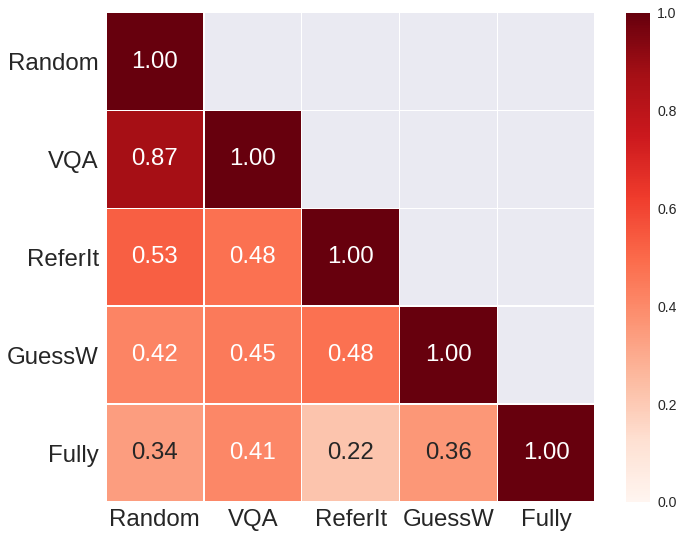}
\caption{RSA scores indicating degree of structural similarity between the multimodal semantic spaces produced by
  the various encoders when receiving 5,000 data points from the FOIL test set consisting of unique images paired with their original captions.} \label{fig:rsa}
\end{figure}

The outputs of the encoders are compared when the same set of inputs is
 given.  We give as input 5,000 data points from the FOIL test set,
 randomly sampled from only the ones with original captions and
 containing unique images, and compare the representations produced by
 the encoders under investigation.
Figure~\ref{fig:rsa} shows that the semantic
space produced by the encoder fully trained on FOIL is rather different from 
all the other models, and that the VQA semantic space is very similar to the one
produced by the randomly initialized encoder.

\paragraph{Nearest neighbour overlap}

We analyse the encoder representations using nearest neighbour
overlap. \cite{collell2018neural} proposed this measure to compare 
the structure of functions that map concepts from
an input to a target space. It is defined as
the number of $k$ nearest neighbours that two paired vectors share in
their respective semantic space. For instance, if $k =3$ and the 3
nearest neighbours of the vector for `cat' $v_{cat}$ in space $V$ are
$\{v_{dog}, v_{tiger}, v_{lion}\}$, and those of the vector of `cat'
$z_{cat}$ in
 space $Z$ are $\{v_{mouse}, v_{tiger}, v_{lion}\}$, the nearest neighbour overlap (NN) is
2. The value is then normalized with respect to the number
of data points and the number of $k$ nearest neighbours. 
 
\begin{table}[h]
\centering
\begin{tabular}{l|cc|cc} \toprule
& \multicolumn{2}{c|}{ $k=1$ } & \multicolumn{2}{c}{$k=10$} \\ 
 & ResNet152 & USE & ResNet152 & USE \\ \midrule
Random  & 0.829 & 0.363 & 0.876 & 0.365\\ 
VQA  & 0.638 & 0.350 & 0.703 & 0.386\\ 
ReferIt  & 0.754 & 0.346 & 0.780 & 0.366\\ 
GuessWhat  & 0.658 & 0.329 & 0.689 & 0.359\\ 
Fully FOIL  & 0.171 & 0.254 & 0.246 & 0.291\\ \bottomrule
\end{tabular}
\caption{Average nearest neighbour overlap between the encoder multimodal
  representations and the ResNet152 and USE embeddings, respectively.}
\label{tab:neighbor}
\end{table}

We take the encoder to be a mapping function from each of the modality-specific representations 
to the multimodal space, and we use the NN 
measure to investigate whether the structure of the multimodal space
produced by the encoder is closer to the visual ResNet152 embeddings
or to the linguistic USE embeddings given as input. We use simple visual and
language inputs, namely, objects and the word corresponding to their object category.  
We consider the 80 object categories of MS-COCO (e.g., dog, car, etc.) and
obtain their USE representations. We build their visual ResNet152
embedding by selecting 100 images for each category from MS-COCO, and
then compute their average. We compute the NN by setting $k=1$ and
 $k=10$.  The results, given in Table~\ref{tab:neighbor}, show that the multimodal spaces learned by all
 the models (except the model with the encoder fully trained on the FOIL task)
 are much closer to the visual input space than to the linguistic
 one. This behaviour could be related to the
   different density of the visual and linguistic semantic spaces of the input data we pointed out in Section~\ref{sec:shared}, 
   where we observed that input images have higher average cosine similarity than input questions, descriptions, and dialogues, respectively.

\section{Conclusion}
\label{sec:conclusion}

Our goal in this paper has been to evaluate the quality of the multimodal representations
learned by an encoder---the core module of all the multimodal models
used currently within the language and vision community---which resembles the cognitive representational hub described by~\cite{patt:theh15}.
Furthermore,
we investigated the transfer potential of the encoded skills, 
taking into account the amount of time (learning epochs) and training
data the models need to adapt to a new task and with how much confidence they make their decisions.
We studied three %
multimodal tasks, where the encoder is trained to 
answer a question about an image (VQA), pick up the object in an image
referred to by a description (ReferIt), and identify the object in an
image that is the target of a goal-oriented question-answer dialogue
(GuessWhat). To carry out this analysis, we have evaluated how the
pre-trained models perform on a diagnostic task, FOIL~\citep{shekhar2017foil_acl}, designed to
check the model's ability to detect semantic incongruence in visually
grounded language. 

Overall, we found that none of the three tasks under investigation leads to learning fine-grained multimodal understanding skills that can solve the FOIL task, although there are differences among tasks.
Our analysis shows that the VQA task is easier to learn (the model
achieves a rather high mean rank precision). However the multimodal encoding skills it
learns are less stable and transferable than the ones learned through
the ReferIt and GuessWhat tasks. This can be seen by the large amount of
data the model has to be exposed to in order to learn the FOIL classification task and by
the unstable results over training epochs. None of the
models transfers their encoding skills with high confidence, but again the VQA
model does it to a lower extent. 

The RSA analysis confirms the higher similarity of the multimodal
spaces generated by the ReferIt and GuessWhat encoders and the high
similarity between the VQA space and the space produced by the randomly initialized 
encoder. From the NN analysis, it appears that for all models (except for the
one fully trained on the FOIL task) the visual modality has higher weight than the
linguistic one in the construction of the multimodal representations.

These differences among tasks %
could be due to subtle parallelisms with the diagnostic task: 
ReferIt and GuessWhat may resemble some aspects of FOIL, since 
 these three tasks revolve around objects (the foiled word is always a noun), while
 arguably the VQA task is more diverse as it contains questions about, e.g., actions,
 attributes, or scene configurations. 
 In future work, it would be interesting to evaluate the models
 on different diagnostic datasets that prioritise skills other than object identification.

\section*{Acknowledgements}
We kindly acknowledge the Leibniz-Zentrum f\"ur Informatik, Dagstuhl
 Seminar 19021 on {\em Joint Processing of Language and Visual Data for Better Automated Understanding}. 
 The Amsterdam team was partially funded by the Netherlands Organisation for Scientific Research (NWO) under VIDI grant nr.~276-89-008, {\em Asymmetry in Conversation}. We gratefully acknowledge the support of NVIDIA
 Corporation with the donation to the University of Trento of the
 GPUs used in our research.

\bibliography{ravi,raffa,raq}

\begin{thebibliography}{}

\bibitem[\protect\citeauthoryear{Adi, Kermany, Belinkov, Lavi, and
  Goldberg}{Adi et~al.}{2017}]{adi2016fine}
Adi, Y., E.~Kermany, Y.~Belinkov, O.~Lavi, and Y.~Goldberg (2017).
\newblock {Fine-grained Analysis of Sentence Embeddings Using Auxiliary
  Prediction Tasks}.
\newblock In {\em International Conference on Learning Representations (ICLR)}.

\bibitem[\protect\citeauthoryear{Alishahi, Barking, and Chrupa{\l}a}{Alishahi
  et~al.}{2017}]{alishahi-etl-conll2017}
Alishahi, A., M.~Barking, and G.~Chrupa{\l}a (2017).
\newblock Encoding of phonology in a recurrent neural model of grounded speech.
\newblock In {\em Proceedings of the 21st Conference on Computational Natural
  Language Learning (CoNLL 2017)}, pp.\  368--378. Association for
  Computational Linguistics.

\bibitem[\protect\citeauthoryear{Antol, Agrawal, Lu, Mitchell, Batra, Zitnick,
  and Parikh}{Antol et~al.}{2015}]{anto:vqa15}
Antol, S., A.~Agrawal, J.~Lu, M.~Mitchell, D.~Batra, C.~L. Zitnick, and
  D.~Parikh (2015).
\newblock {VQA}: Visual question answering.
\newblock In {\em International Conference on Computer Vision (ICCV)}.

\bibitem[\protect\citeauthoryear{Barrault, Bougares, Specia, Lala, Elliott, and
  Frank}{Barrault et~al.}{2018}]{barr:find18}
Barrault, L., F.~Bougares, L.~Specia, C.~Lala, D.~Elliott, and S.~Frank (2018).
\newblock Findings of the third shared task on multimodal machine translation.
\newblock In {\em Proceedings of the Third Conference on Machine Translation
  (WMT)}, Volume~2, pp.\  304–323. Association for Computational Linguistics.

\bibitem[\protect\citeauthoryear{Bouchacourt and Baroni}{Bouchacourt and
  Baroni}{2018}]{bouchacourt2018agents}
Bouchacourt, D. and M.~Baroni (2018).
\newblock How agents see things: On visual representations in an emergent
  language game.
\newblock In {\em Proceedings of the 2018 Conference on Empirical Methods in
  Natural Language Processing}, pp.\  981--985. Association for Computational
  Linguistics.

\bibitem[\protect\citeauthoryear{Cer, Yang, Kong, Hua, Limtiaco, John,
  Constant, Guajardo-Cespedes, Yuan, Tar, et~al.}{Cer
  et~al.}{2018}]{cer2018universal}
Cer, D., Y.~Yang, S.-y. Kong, N.~Hua, N.~Limtiaco, R.~S. John, N.~Constant,
  M.~Guajardo-Cespedes, S.~Yuan, C.~Tar, et~al. (2018).
\newblock Universal sentence encoder.
\newblock {\em arXiv preprint arXiv:1803.11175\/}.

\bibitem[\protect\citeauthoryear{Collell and Moens}{Collell and
  Moens}{2018}]{collell2018neural}
Collell, G. and M.-F. Moens (2018).
\newblock Do neural network cross-modal mappings really bridge modalities?
\newblock {\em arXiv preprint arXiv:1805.07616\/}.

\bibitem[\protect\citeauthoryear{Conneau, Kiela, Schwenk, Barrault, and
  Bordes}{Conneau et~al.}{2017}]{cann:super17}
Conneau, A., D.~Kiela, H.~Schwenk, L.~Barrault, and A.~Bordes (2017).
\newblock Supervised learning of universal sentence representations from
  natural language inference data.
\newblock In {\em Proceedings of the 2017 Conference on Empirical Methods in
  Natural Language Processing}, pp.\  670--680.

\bibitem[\protect\citeauthoryear{Conneau, Kruszewski, Lampl, Barrault, and
  Baroni}{Conneau et~al.}{2018}]{conn:what18}
Conneau, A., G.~Kruszewski, G.~Lampl, L.~Barrault, and M.~Baroni (2018).
\newblock What you can cram into a single {$\backslash$}{\$}{\&}!{\#}* vector:
  Probing sentence embeddings for linguistic properties.
\newblock In {\em Proceedings of ACL}.

\bibitem[\protect\citeauthoryear{Das, Kottur, Gupta, Singh, Yadav, Moura,
  Parikh, and Batra}{Das et~al.}{2017}]{visdial}
Das, A., S.~Kottur, K.~Gupta, A.~Singh, D.~Yadav, J.~M. Moura, D.~Parikh, and
  D.~Batra (2017).
\newblock {V}isual {D}ialog.
\newblock In {\em Proceedings of the IEEE Conference on Computer Vision and
  Pattern Recognition}.

\bibitem[\protect\citeauthoryear{de~Vries, Strub, Chandar, Pietquin,
  Larochelle, and Courville}{de~Vries et~al.}{2017}]{guesswhat_game}
de~Vries, H., F.~Strub, S.~Chandar, O.~Pietquin, H.~Larochelle, and A.~C.
  Courville (2017).
\newblock Guesswhat?! {V}isual object discovery through multi-modal dialogue.
\newblock In {\em Conference on Computer Vision and Pattern Recognition
  (CVPR)}.

\bibitem[\protect\citeauthoryear{Dobnik, Ghanimifard, and Kelleher}{Dobnik
  et~al.}{2018}]{dobn:expl18}
Dobnik, S., M.~Ghanimifard, and J.~D. Kelleher (2018).
\newblock Exploring the functional and geometric bias of spatial relations
  using neural language models.
\newblock In {\em Proceedings of the First International Workshop on Spatial
  Language Understanding (SpLu)}, pp.\  1--11.

\bibitem[\protect\citeauthoryear{He, Zhang, Ren, and Sun}{He
  et~al.}{2016}]{he2016:resnet}
He, K., X.~Zhang, S.~Ren, and J.~Sun (2016).
\newblock Deep residual learning for image recognition.
\newblock In {\em Proceedings of the IEEE conference on computer vision and
  pattern recognition}, pp.\  770--778.

\bibitem[\protect\citeauthoryear{Johnson, Hariharan, van~der Maaten, Fei-Fei,
  Zitnick, and Girshick}{Johnson et~al.}{2017}]{girs:clev16}
Johnson, J., B.~Hariharan, L.~van~der Maaten, L.~Fei-Fei, C.~L. Zitnick, and
  R.~Girshick (2017).
\newblock Clevr: A diagnostic dataset for compositional language and elementary
  visual reasoning.
\newblock In {\em Proceedings of CVPR 2017}.

\bibitem[\protect\citeauthoryear{Kazemzadeh, Ordonez, Matten, and
  Berg}{Kazemzadeh et~al.}{2014}]{KazemzadehOrdonezMattenBergEMNLP14}
Kazemzadeh, S., V.~Ordonez, M.~Matten, and T.~L. Berg (2014).
\newblock Referit game: Referring to objects in photographs of natural scenes.
\newblock In {\em EMNLP}.

\bibitem[\protect\citeauthoryear{Kingma and Ba}{Kingma and
  Ba}{2014}]{kingma2014:adam}
Kingma, D.~P. and J.~Ba (2014).
\newblock Adam: A method for stochastic optimization.
\newblock {\em arXiv preprint arXiv:1412.6980\/}.

\bibitem[\protect\citeauthoryear{Kriegeskorte, Mur, and
  Bandettini}{Kriegeskorte et~al.}{2008}]{kriegeskorte2008representational}
Kriegeskorte, N., M.~Mur, and P.~A. Bandettini (2008).
\newblock Representational similarity analysis-connecting the branches of
  systems neuroscience.
\newblock {\em Frontiers in systems neuroscience\/}~{\em 2}, 4.

\bibitem[\protect\citeauthoryear{Lin, Maire, Belongie, Hays, Perona, Ramanan,
  Dollar, P., and Zitnick}{Lin et~al.}{2014}]{lin:micr14}
Lin, T.-Y., M.~Maire, S.~Belongie, J.~Hays, P.~Perona, D.~Ramanan, Dollar, P.,
  and C.~L. Zitnick (2014).
\newblock Microsoft {COCO}: Common objects in context.
\newblock In {\em Proceedings of ECCV (European Conference on Computer
  Vision)}.

\bibitem[\protect\citeauthoryear{Linzen, Dupoux, and Goldberg}{Linzen
  et~al.}{2016}]{linzen2016assessing}
Linzen, T., E.~Dupoux, and Y.~Goldberg (2016).
\newblock {Assessing the Ability of LSTMs to Learn Syntax-Sensitive
  Dependencies}.
\newblock {\em Transactions of the Association for Computational
  Linguistics\/}~{\em 4}, 521--535.

\bibitem[\protect\citeauthoryear{Madhysastha, Wang, and Specia}{Madhysastha
  et~al.}{2018}]{madhysastha-wang-specia_NAACL:2018}
Madhysastha, P., J.~Wang, and L.~Specia (2018).
\newblock Defoiling foiled image captions.
\newblock In {\em Conference of the North American Chapter of the Association
  for Computational Linguistics: Human Language Technologies}, New Orleans, LA.

\bibitem[\protect\citeauthoryear{Mostafazadeh, Misra, Devlin, Mitchell, He, and
  Vanderwende}{Mostafazadeh et~al.}{2016}]{gene:most16}
Mostafazadeh, N., I.~Misra, J.~Devlin, M.~Mitchell, X.~He, and L.~Vanderwende
  (2016, August).
\newblock Generating natural questions about an image.
\newblock In {\em Proceedings of the 54th Annual Meeting of the Association for
  Computational Linguistics (Volume 1: Long Papers)}, Berlin, Germany, pp.\
  1802--1813. Association for Computational Linguistics.

\bibitem[\protect\citeauthoryear{Patterson and Ralph}{Patterson and
  Ralph}{2015}]{patt:theh15}
Patterson, K. and M.~A.~L. Ralph (2015).
\newblock {\em Neurobiology of Language}, Chapter The Hub-and-Spoke Hypothesis
  of Semantic Memory.
\newblock Elsevier.

\bibitem[\protect\citeauthoryear{Razavian, Azizpour, Sullivan, and
  Carlsson}{Razavian et~al.}{2014}]{raza:Cnnf14}
Razavian, A.~S., H.~Azizpour, J.~Sullivan, and S.~Carlsson (2014).
\newblock {CNN} features off-the-shelf: an astounding baseline for recognition.
\newblock In {\em Proceedings of the IEEE Conference on Computer Vision and
  Pattern Recognition Workshops}, pp.\  806--813.

\bibitem[\protect\citeauthoryear{Shekhar, Pezzelle, Klimovich, Herbelot, Nabi,
  Sangineto, and Bernardi}{Shekhar et~al.}{2017}]{shekhar2017foil_acl}
Shekhar, R., S.~Pezzelle, Y.~Klimovich, A.~Herbelot, M.~Nabi, E.~Sangineto, and
  R.~Bernardi (2017).
\newblock "foil it! find one mismatch between image and language caption".
\newblock In {\em Proceedings of the 55th Annual Meeting of the Association for
  Computational Linguistics (ACL) (Volume 1: Long Papers)}, pp.\  255--265.

\bibitem[\protect\citeauthoryear{Suhr, Lewis, Yeh, and Artzi}{Suhr
  et~al.}{2017}]{suhr-EtAl:2017:Short}
Suhr, A., M.~Lewis, J.~Yeh, and Y.~Artzi (2017, July).
\newblock A corpus of natural language for visual reasoning.
\newblock In {\em Proceedings of the Annual Meeting of the Association for
  Computational Linguistics}, Vancouver, Canada, pp.\  217--223. Association
  for Computational Linguistics.

\bibitem[\protect\citeauthoryear{Wieting and Kiela}{Wieting and
  Kiela}{2019}]{wieting2019no}
Wieting, J. and D.~Kiela (2019).
\newblock No training required: Exploring random encoders for sentence
  classification.
\newblock In {\em ICLR (accepted)}.

\bibitem[\protect\citeauthoryear{Yu, Poirson, Yang, Berg, and Berg}{Yu
  et~al.}{2016}]{yu2016modeling}
Yu, L., P.~Poirson, S.~Yang, A.~C. Berg, and T.~L. Berg (2016).
\newblock Modeling context in referring expressions.
\newblock In {\em European Conference on Computer Vision}, pp.\  69--85.
  Springer.

\bibitem[\protect\citeauthoryear{Zhang and Bowman}{Zhang and
  Bowman}{2018}]{zhan:lang18}
Zhang, K.~W. and S.~R. Bowman (2018).
\newblock Language modeling teaches you more syntax than translation does:
  Lessons learned through auxiliary task analysis.
\newblock In {\em Proceedings of the 2018 EMNLP Workshop BlackboxNLP: Analyzing
  and Interpreting Neural Networks for NLP}, pp.\  359--361. Association for
  Computational Linguistics.

\end{thebibliography}
\bibliographystyle{chicago} %

\end{document}